\documentclass[runningheads]{llncs}

\usepackage{eccv}

\usepackage{eccvabbrv}

\usepackage[utf8]{inputenc} 
\usepackage[T1]{fontenc}    
\usepackage{url}            
\usepackage{booktabs}       
\usepackage{amsfonts}       
\usepackage{nicefrac}       
\usepackage{microtype}      
\usepackage{xcolor}         
\usepackage{graphicx}
\usepackage{overpic}
\usepackage{multirow}
\usepackage{colortbl}
\usepackage{caption}
\usepackage{amssymb}
\usepackage{bbding}
\usepackage{utfsym}
\usepackage{amsmath}

\usepackage[accsupp]{axessibility}

\newcommand{\figref}[1]{Fig.~\ref{#1}}
\newcommand{\tabref}[1]{Tab.~\ref{#1}}

\newcommand{\secref}[1]{Sec.~\ref{#1}}
\newcommand{\smallsec}[1]{\vspace{3pt}\noindent\textbf{#1}}

\usepackage{hyperref}

\usepackage{orcidlink}

\begin{document}

\title{FlowPainter: Inpainting Optical Flow\\ via Confidence-Guided Completion} 

\titlerunning{FlowPainter}

\author{Yuang Meng\inst{1,3} \and
Chenyang Wu\inst{1} \and
Xianshun Liu\inst{1,3} \and
Chun-Le Guo\inst{1} \thanks{C. L. Guo is the corresponding author.} \and \\
Zichen Liang\inst{1} \and 
Lina Lei\inst{1} \and 
Jie Liang\inst{3} \and 
Hui Zeng\inst{3} \and 
Chongyi Li\inst{1} \and 
Lei Zhang\inst{2,3}}

\authorrunning{Y.~Meng et al.}

\institute{VCIP, CS, Nankai University \and
The Hong Kong Polytechnic University \and
OPPO Research Institute \\
\email{mya@mail.nankai.edu.cn}
}

\maketitle

\begin{abstract}
Existing optical flow estimation methods broadly follow two paradigms: iterative optimization and diffusion-based estimation.
Iterative methods, exemplified by RAFT, achieve accurate flow estimation through recurrent refinement, but can still be challenged by large displacements and complex motion patterns.
Diffusion-based methods introduce generative modeling into optical flow and have shown promising results in such ambiguous regions.
However, existing diffusion-based flow models usually denoise the entire dense flow field from Gaussian noise, including simple regions where reliable motion structure can already be estimated by a lightweight network.
This increases the denoising burden and may lead to slow convergence and unstable training.
To address these issues, we introduce FlowPainter, a diffusion-based optical flow framework that reformulates dense-flow generation as confidence-guided soft inpainting.
FlowPainter first employs a lightweight confidence-aware network to predict a rough flow and a pixel-wise confidence mask, which serves as a reliability gate for distinguishing reliable simple regions from uncertain hard regions.
The resulting simple-flow prior is used for confidence-based initialization and is further injected into the iterative denoising process through confidence-gated residual guidance.
With a dynamically decaying guidance strength, FlowPainter stabilizes early denoising while preserving the flexibility of the diffusion model for late-stage detail refinement.
Extensive experiments on public benchmarks, including Sintel, KITTI, and Spring, demonstrate that FlowPainter achieves strong accuracy under comparable training settings and improves convergence efficiency over existing diffusion-based optical flow methods, with notable gains on challenging benchmark splits.
Our approach provides a practical direction for integrating reliable discriminative priors with diffusion-based refinement for optical flow estimation. 
Our code is made publicly available at \href{https://github.com/mya012/FlowPainter}{https://github.com/mya012/FlowPainter}.
\end{abstract}

\section{Introduction}
\label{sec:intro}

Optical flow estimation is a fundamental problem in computer vision, aiming to infer a dense motion vector for each pixel between consecutive frames. 
It is widely used in video understanding \cite{liu2025flow4agent,fan2018end}, action recognition \cite{piergiovanni2019representation,sun2018optical}, autonomous driving \cite{mahjourian2022occupancy,shen2023optical,capito2020optical}, video frame interpolation \cite{raket2012motion,wu2022video,hai2025hierarchical}, and restoration problems \cite{chen2026calibrated,meng2025ultraled,qu2026there}. 
Despite decades of progress, accurate optical flow estimation remains challenging in real-world scenarios due to fast motion, occlusions, illumination changes, and textureless regions.

Deep learning has greatly advanced optical flow estimation.
Early CNN-based methods \cite{dosovitskiy2015flownet,ilg2017flownet,ranjan2017optical,sun2018pwc} introduced end-to-end regression, stacked refinement, and coarse-to-fine matching, but their fixed receptive fields and limited feature expressiveness can hinder complex motion modeling.
RAFT \cite{teed2020raft} established a strong iterative refinement paradigm by building an all-pairs correlation volume and recurrently updating the flow field.
Its follow-up works \cite{jiang2021learning,teed2021raft,wang2024sea,poggi2025flowseek} further improve global aggregation, 3D motion modeling, training strategies, and practical efficiency.
Nevertheless, iterative optimization can still be challenged by very large displacements, severe occlusions, and complex non-rigid motion, where correlation construction and recurrent updates may become less reliable \cite{luo2024flowdiffuser}.

In parallel, diffusion models have recently been explored for optical flow estimation \cite{luo2024flowdiffuser,saxena2023surprising,dong2023open,pepe2025geometric}.
Methods such as DDVM \cite{saxena2023surprising} and FlowDiffuser \cite{luo2024flowdiffuser} formulate optical flow estimation as a denoising process from Gaussian noise to dense flow, conditioned on image features.
By introducing generative modeling into flow estimation, these methods show promising results in ambiguous regions with large displacement or complex motion.
However, existing diffusion-based flow models usually denoise the whole dense flow field from noise, including simple regions where a lightweight estimator can already provide reliable motion structure.
This increases the denoising burden and may slow convergence or destabilize training, especially when the model has to reconstruct precise flow solely from Gaussian noise.

\begin{figure}[h]
  \centering
  \includegraphics[height=4.7cm]{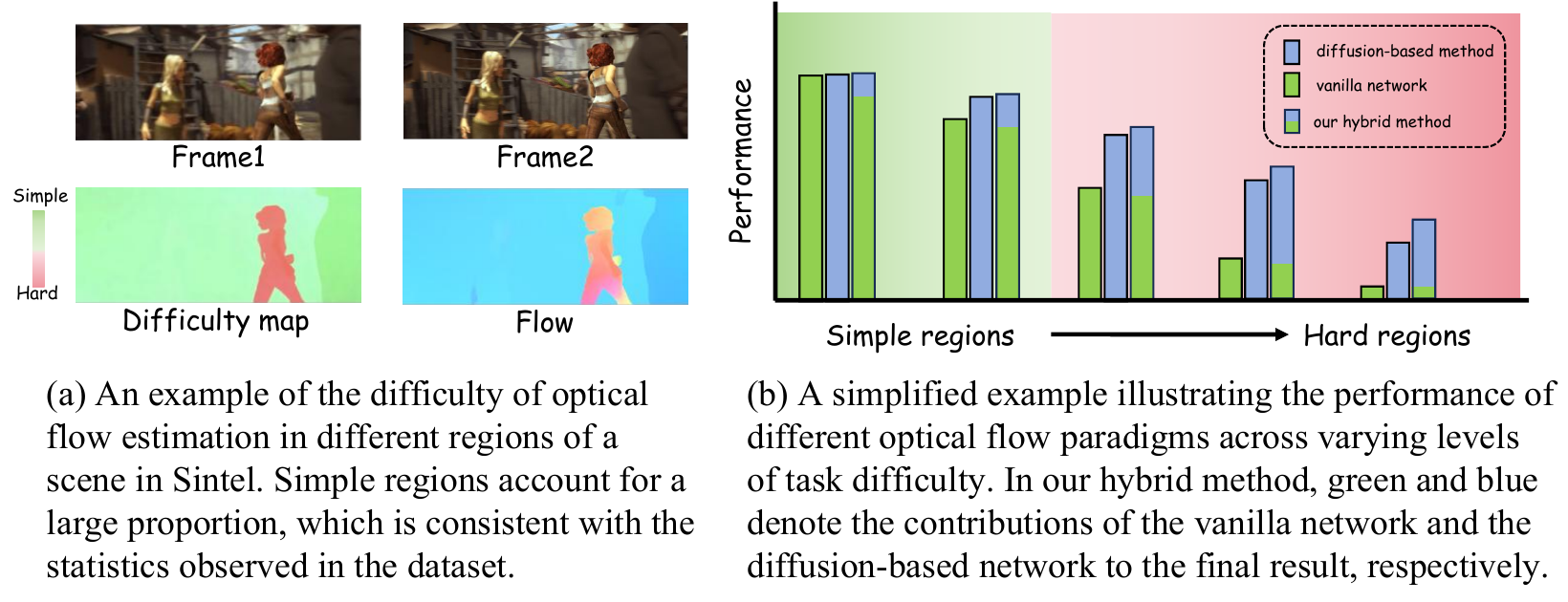}
  \caption{Illustration of region-wise flow difficulty in real scenes and the resulting motivation for our hybrid paradigm across different optical flow methods. 
  }
  \label{fig:intro_fig}
\end{figure}

A closer examination of the two paradigms, i.e., iterative optimization \cite{jiang2021learning,teed2021raft,wang2024sea,poggi2025flowseek} and diffusion-based approaches \cite{luo2024flowdiffuser,saxena2023surprising,pepe2025geometric}, suggests an important observation. 
As shown in \figref{fig:intro_fig} (a), the example depicts a moving person and a background that shifts in tandem. The rapid motion of the person and the non-rigid deformation of the human body make optical flow estimation challenging in the foreground. In contrast, estimating the flow of the background regions that translate alongside the person is relatively simpler. This observation corresponds to the trend illustrated in \figref{fig:intro_fig} (b). 
For "simple regions" with small motion, usually clear textures or smooth backgrounds, even a vanilla network in the iterative optimization paradigm can produce flow estimates comparable to those of a diffusion-based method. 
In contrast, for "hard regions" involving large displacements, occlusions, or motion boundaries, such a vanilla network often struggles, whereas diffusion models can leverage their generative modeling ability to refine or recover plausible motion.
Meanwhile, diffusion-based methods usually apply the denoising process to the whole flow field, including large simple regions, which increases the denoising burden for regions where reliable motion structure is already available.
We provide further empirical evidence for this viewpoint in \secref{sec:motivation}.
This motivates a hybrid strategy that combines the strengths of both paradigms: using a vanilla network to efficiently handle simple regions and provide informative priors, while guiding a diffusion model to refine uncertain hard regions.
In this way, the diffusion model can better exploit its refinement ability in hard-flow regions, while the vanilla network provides structural priors that help stabilize the generation process and improve convergence efficiency.

\begin{figure}[h]
  \centering
  \includegraphics[height=4.5cm]{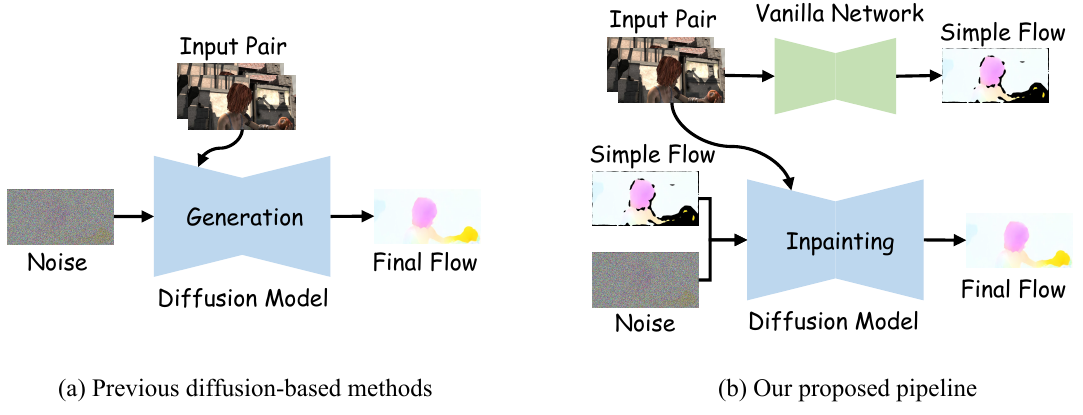}
  \caption{An overview of previous diffusion-based methods and our proposed framework \textbf{FlowPainter}. Previous diffusion-based methods generated optical flow entirely from random Gaussian noise. FlowPainter first employs a vanilla network to generate simple optical flow for small motion regions, then uses this as prior information to guide the inpainting of the final optical flow.
  }
  \label{fig:teaser}
\end{figure}

Based on this idea, we propose \textbf{FlowPainter}, a confidence-guided diffusion framework for optical flow estimation, as shown in \figref{fig:teaser}.
FlowPainter first employs a lightweight Confidence-Aware Network to predict a rough flow and a pixel-wise confidence mask.
The confidence mask serves as a reliability gate: high-confidence regions provide simple-flow priors, while low-confidence regions rely more on image conditioning and diffusion refinement.
The simple-flow prior is used for confidence-based initialization and is further injected into the denoising process through confidence-gated residual guidance.
A time-decayed guidance schedule stabilizes early denoising and gradually relaxes the prior constraint for late-stage detail refinement.
Together with standard image conditioning features used in diffusion-based optical flow \cite{luo2024flowdiffuser}, this design reformulates dense-flow generation as a soft inpainting-style refinement problem, improving training stability and final accuracy over existing diffusion-based optical flow methods.

Our main contributions are summarized as follows:

\begin{itemize}
\item We propose a confidence-guided optical flow framework that integrates a lightweight Confidence-Aware Network with diffusion-based refinement, using reliable rough flow as a structural prior for soft inpainting-style denoising.
\item We introduce confidence-based initialization and confidence-gated residual guidance with a time-decayed schedule, reducing the denoising burden in reliable simple regions while preserving flexibility for uncertain hard regions.
\item Extensive experiments on Sintel \cite{Butler:ECCV:2012}, KITTI \cite{Menze2015CVPR}, and Spring \cite{Mehl2023_Spring} show that FlowPainter achieves strong accuracy under comparable training settings and improves convergence efficiency over existing diffusion-based optical flow methods.
\end{itemize}

\section{Related Work}
\label{sec:relate_work}

\subsection{Iterative Optimization Paradigm for Optical Flow}

End-to-end deep optical flow estimation started with FlowNet \cite{dosovitskiy2015flownet}, which first showed that convolutional neural networks can directly regress dense flow from image pairs. 
FlowNet2 \cite{ilg2017flownet} improved accuracy via stacking and specialized small-displacement branches, at the cost of higher computation and memory. 
SPyNet \cite{ranjan2017optical} adopted a coarse-to-fine pyramid to better handle large motions, while PWC-Net \cite{sun2018pwc} combined pyramid features, warping and a cost volume to achieve a stronger accuracy–efficiency trade-off.

RAFT \cite{teed2020raft} established a highly influential iterative refinement paradigm by building a high-resolution 4D all-pairs correlation volume and repeatedly updating the flow with a recurrent module. Subsequent work \cite{jiang2021learning,teed2021raft,xu2022gmflow,wang2024sea,poggi2025flowseek,Zhang2021SepFlow} largely advances along stronger correlation representations and more effective update. 
GMA \cite{jiang2021learning} enhances matching and updates under occlusions and long-range dependencies via global motion aggregation, and RAFT-3D \cite{teed2021raft} extends similar ideas to 3D scene flow. 
In parallel, approaches with stronger global modeling mitigate the limitations of local matching for extreme displacements: GMFlow \cite{xu2022gmflow} leverages Transformer-style global matching, and FlowFormer \cite{huang2022flowformer} tokenizes cost volumes and uses Transformer memory for global aggregation and decoding, showing strong performance on datasets such as Sintel \cite{Wulff:ECCVws:2012,Butler:ECCV:2012}.

More recently, SEA-RAFT \cite{wang2024sea} preserves the iterative refinement backbone while simplifying and strengthening loss function and pretraining strategies to improve accuracy–speed trade-offs. 
For resource-constrained settings, compact designs such as FlowSeek \cite{poggi2025flowseek} reduce training and deployment costs by using lightweight architectures and external priors (e.g., foundation-model knowledge or low-dimensional motion parameterizations), improving practical usability. 
Overall, iterative optimization \cite{teed2020raft,jiang2021learning,xu2022gmflow,wang2024sea,poggi2025flowseek} is highly effective for progressive correction, yet it can still be limited when motion is extreme, occlusions are severe or non-rigid deformation is complex, where matching errors may be amplified. 
Moreover, the update mechanism remains largely discriminative, lacking explicit uncertainty modeling in ambiguous regions \cite{wang2024sea, luo2024flowdiffuser}—motivating exploration of generative paradigms such as diffusion.

\subsection{Diffusion Models and Applications}

Diffusion models have rapidly progressed from high-fidelity synthesis to broad conditional generation \cite{zhang2023adding,zhang2024clay,wang2025diffusion,zhao2026resilphaseplugandplayphasemapping} and inverse problem solving \cite{gao2023implicit,wang2024exploiting}. 
DDPM \cite{ho2020denoising} established the standard formulation of progressive noising and iterative denoising; DDIM \cite{song2020denoising} introduced implicit sampling trajectories that substantially accelerate inference while remaining compatible with DDPM training. 

Motivated by their generative capacity and uncertainty representation, diffusion models have been applied to dense prediction \cite{ji2023ddp,saxena2023surprising,luo2024flowdiffuser}, including optical flow. 
DDVM \cite{saxena2023surprising} first formulated flow estimation as conditional generation of flow fields from an image pair, demonstrating competitive performance but often requiring heavy pretraining or substantial computing resources. 
Reproduction efforts such as Open-DDVM \cite{dong2023open} highlight that injecting structural priors such as multi-scale correlation volumes can significantly improve convergence and performance. 
FlowDiffuser \cite{luo2024flowdiffuser} further explores combining diffusion denoising with recurrent decoding and historical states to improve efficiency and better exploit iterative information, while GA-DDVM \cite{pepe2025geometric} introduces geometric priors via geometric algebra constraints, underscoring the effectiveness of embedding task structure into the diffusion process.

Despite promising results for large displacements, occlusions and complex motion, existing diffusion-based optical flow \cite{luo2024flowdiffuser,saxena2023surprising,dong2023open,pepe2025geometric} faces two key limitations. 
First, these approaches frequently tend to spend a significant amount of computational cost on simple regions, which limits the computation that can be allocated to truly challenging areas. 
Second, they are challenged by the difficulty of generating dense flow from Gaussian noise, which slows convergence and can destabilize training—especially in early denoising stages where reliable intermediate structure is lacking. 
Consequently, a key practical direction is to introduce controllable and reliable priors that accelerate and stabilize optimization without weakening diffusion’s expressive power, which naturally motivates our confidence-guided diffusion framework.

\section{Methodology}
\label{sec:method}

\subsection{Motivation}
\label{sec:motivation}

For optical flow estimation, we first introduce a practical working partition of motion patterns. 
We refer to the flow in small-displacement or otherwise easy-to-model regions as \textbf{simple flow}, and denote the flow in large-displacement or highly complex motion regions as \textbf{hard flow}. 
It is worth noting that this partition is not intended to provide a complete definition of flow difficulty. Instead, flow magnitude is used as a practical proxy for large-displacement difficulty, while other factors such as occlusion, texture ambiguity, motion boundaries, and non-rigid deformation can also make optical flow estimation challenging.

\smallsec{Identification and Experiments of Issues with Existing Methods. }
Although diffusion-based optical flow methods \cite{saxena2023surprising, luo2024flowdiffuser} have shown noticeable improvements over traditional estimators, we observe that these gains are largely concentrated in scenarios containing substantial hard flow. 
In other words, when a scene is dominated by simple flow, the advantage of diffusion-based approaches largely diminishes, while their iterative sampling still incurs significantly higher computational cost. 
In real-world scenes, however, simple and hard flow typically coexist. 
For such mixed-motion scenes, the benefit of diffusion-based methods again mainly comes from regions with large and complex motion. 
We support this observation experimentally on the Sintel(train)  \cite{Butler:ECCV:2012} benchmark by comparing FlowDiffuser \cite{luo2024flowdiffuser} with a vanilla network employing an architecture similar to PWC-Net \cite{sun2018pwc}. 
Concretely, when computing evaluation metrics, we restrict the computation to pixels whose ground-truth flow magnitude is below a threshold. 
We vary this threshold over {5, 10, 20, 40, $\infty$} (with $\infty$ indicating no thresholding). 
As summarized in \tabref{tab:Value_Compare}, the observed performance gap in favor of diffusion-based estimation \cite{luo2024flowdiffuser} consistently increases as the threshold becomes larger, indicating that diffusion models provide progressively greater benefits as larger-displacement and more challenging pixels are included in the evaluation.

\begin{table}[t]
  \centering
  \caption{Endpoint Error results of FlowDiffuser \cite{luo2024flowdiffuser} and Vanilla Network on the clean/final pass of Sintel \cite{Butler:ECCV:2012} dataset under different optical flow threshold constraints.}
  \setlength{\tabcolsep}{3pt}
    \begin{tabular}{ccccccc}
    \toprule
\cmidrule{1-7}    \multicolumn{2}{c}{Threshold} & 5     & 10    & 20    & 40    & $\infty$ \\
    \midrule
    \multicolumn{2}{c}{FlowDiffuser \cite{luo2024flowdiffuser}} & 0.121/0.281 & 0.158/0.384 & 0.225/0.639 & 0.323/0.916 & 0.877/2.271 \\
    \multicolumn{2}{c}{Vanilla Network} & 0.132/0.295 & 0.176/0.415 & 0.255/0.750 & 0.421/1.119 & 1.545/3.849 \\
    \bottomrule
    \end{tabular}%
  \label{tab:Value_Compare}%
\end{table}%

\smallsec{Proposal of Hybrid Method. }
Above findings motivate a hybrid perspective on an effective solution for optical flow. 
A vanilla network is highly effective in simple regions, where it achieves strong accuracy at substantially lower cost. 
However, as motion complexity increases, its performance gap with diffusion-based methods becomes more evident, indicating that the two paradigms exhibit complementary strengths. 
At the same time, standard diffusion-based approaches still denoise the whole dense flow field from noise, including regions where a lightweight estimator already provides reliable structure. This increases the denoising burden and makes the generation process unnecessarily difficult in simple regions. 
Moreover, because recovering accurate flow from pure Gaussian noise is inherently difficult, prior diffusion-based flow methods \cite{saxena2023surprising,luo2024flowdiffuser} often suffer from unstable training and slow convergence. 
Therefore, using the reliable predictions of a vanilla network as a prior can reduce the effective difficulty of diffusion-based flow generation, while enabling the diffusion model to bias refinement toward complex or uncertain regions where generative modeling is most beneficial, leading to more stable optimization and improved overall performance.

\subsection{Confidence-Aware Network}
\label{sec:method_stage1}

\begin{figure}[h]
  \centering
  \includegraphics[height=4.9cm]{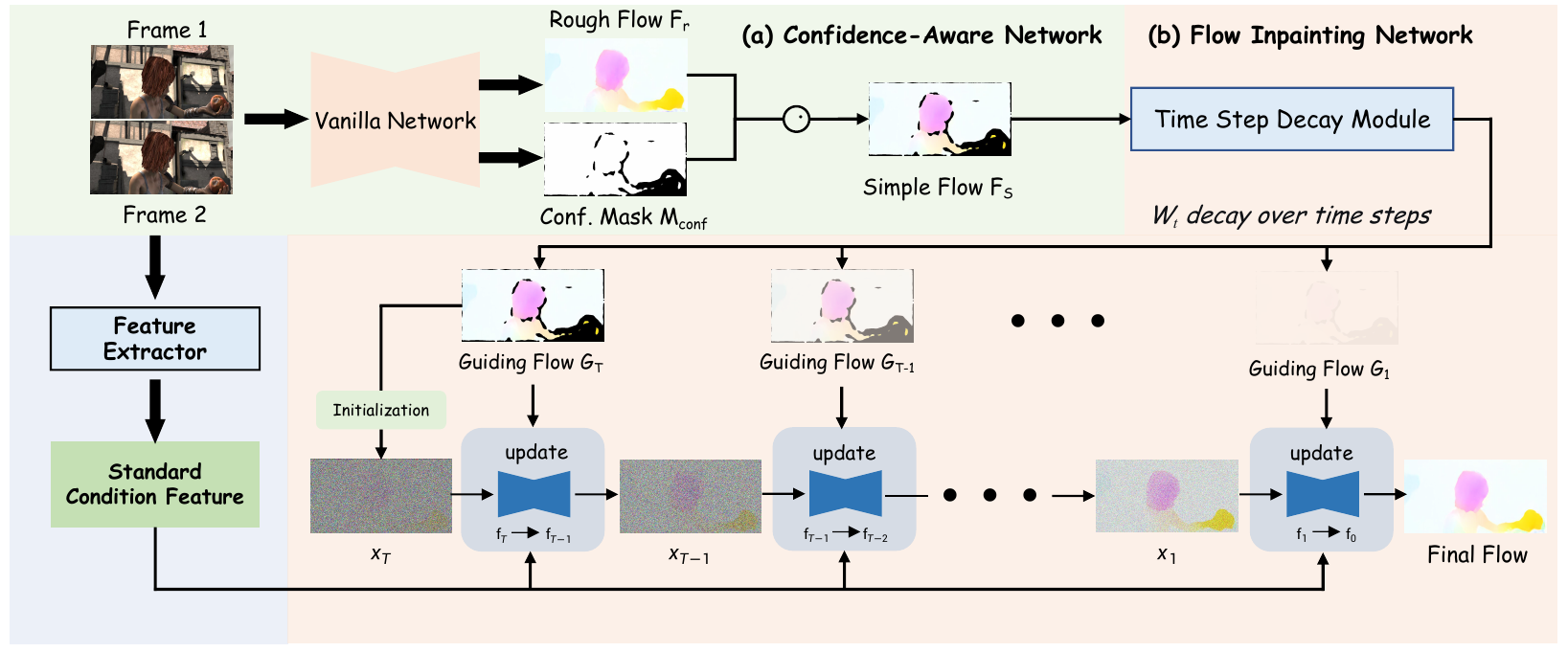}
  \caption{An overview of our proposed framework \textbf{FlowPainter},  which consists of two main components: \textbf{(a) Confidence-Aware Network}: A lightweight PWC-Net-based model \cite{sun2018pwc,zhao2020maskflownet} predicts a rough flow and a pixel-wise confidence mask indicating motion reliability. The confidence mask serves as a reliability gate, providing a simple-flow prior in reliable regions and suppressing unreliable predictions in challenging areas. 
  \textbf{(b) Flow Inpainting Network}: The prior is first used for confidence-based initialization and is then injected as residual guidance through a Time Step Decay Module. The guidance is strong in early denoising steps to stabilize coarse structure, and gradually decays in later steps to preserve the denoiser's flexibility for detail refinement. We also condition denoising on features extracted from input frames to support accurate flow synthesis in challenging regions such as occlusions and motion boundaries.
  }
  \label{fig:method}
\end{figure}

The Confidence-Aware Network is designed to estimate simple flow efficiently, as shown in \figref{fig:method} (a). 
It takes the same input as standard optical flow models—two consecutive frames, but produces two outputs: \textit{1) a rough dense optical flow field $F_{r}$} and \textit{2) a pixel-wise confidence mask $M_{conf}$}. 
Among them, $M_{conf}$ is defined on [0,1], where 0 indicates that $F_{r}$ is entirely unreliable in the corresponding region, and 1 indicates that the corresponding prediction is highly reliable. 
\textit{We refer to the ground truth for $F_{r}$ and $M_{conf}$ during training as $F_{gt}$ and $M_{gt}$, respectively. }

\smallsec{Supervision of Confidence Mask. }
Architecturally, our Confidence-Aware Network follows a PWC-Net-style design \cite{sun2018pwc,zhao2020maskflownet}, but with substantially fewer layers and parameters. 
\textit{The key challenge in training this model lies in how to construct a reliable $M_{gt}$ to supervise $M_{conf}$ during the training phase. } 
Conceptually, $M_{gt}$ is intended to encode two complementary sources of unreliability: hard-flow regions and occluded regions.
We denote the corresponding reliability maps as $M_{hf}$ and $M_{occ}$, respectively.
As discussed in \secref{sec:motivation}, $M_{hf}$ is characterized by large displacements and complex motion patterns. 
Due to its compact capacity and simplified structure, the Confidence-Aware Network typically cannot represent such motions well, and therefore should assign these regions lower confidence. 
$M_{hf}$ is constructed as follows during training:
\begin{equation}
\begin{split}
    M_{d} &= Norm(MSE(F_{gt}, F_{r})), \\
    M_{hf} &= 1 - Er(M_{d}),
\end{split}
\end{equation}
where $MSE$ denotes the mean squared error and $Norm$ represents normalization. $M_{d}$ denotes the normalized difference map, and $Er$ denotes an erosion operation implemented with a $7 \times 7$ kernel to conservatively shrink reliable regions around motion boundaries and uncertain borders.
$M_{occ}$ is defined as a reliability map for occlusions, where visible regions are assigned high confidence and occluded regions are assigned zero confidence. Occluded regions correspond to areas where motion causes correspondences between adjacent frames to be lost, violating the brightness constancy assumption that underlies optical flow computation.
Prior work has shown that occlusions can significantly degrade flow accuracy \cite{zhao2020maskflownet}. Consequently, we explicitly force the predicted confidence values in occluded regions to be zero. We employ forward-backward consistency check \cite{xu2022gmflow} to obtain $M_{occ}$ during the training phase.
In summary, $M_{gt}$ is constructed as follows while training:
\begin{equation}
    M_{gt} = Min(M_{hf}, M_{occ}),
\end{equation}
where $Min$ represents taking the minimum value at each corresponding pixel location. This design ensures that either hard motion or occlusion leads to low confidence, making $M_{conf}$ a reliability gate rather than a magnitude correction term.

\smallsec{General Training Strategy. }
To ensure stable optimization, we adopt a staged training strategy for the Confidence-Aware Network. 
We first train the network to convergence using only optical flow supervision and self-supervised objectives, ensuring that the network's output $F_{r}$ is reasonably accurate. 
After the flow prediction stabilizes, we introduce the confidence-mask supervision and finally train both objectives jointly to convergence. 
Detailed training settings and implementation specifics are provided in \secref{sec:exp_details}.

\subsection{Flow Inpainting Network}
\label{sec:method_stage2}

In our proposed Flow Inpainting Network, optical flow estimation is reformulated as a soft inpainting-style refinement problem conditioned on reliable simple-flow regions, as shown in \figref{fig:method} (b). 
Existing diffusion-based optical flow approaches \cite{saxena2023surprising,luo2024flowdiffuser} rely exclusively on image-derived features extracted from input frames as conditioning signals during denoising. 
This formulation implicitly requires the model to recover both simple and complex motion patterns from pure Gaussian noise, which makes optimization difficult and often leads to unstable training and slow convergence overall. 
Our key idea is to introduce the estimated simple flow as an additional structural prior, so that reliable regions provide useful motion structure and uncertain regions are left to be refined by image conditioning and diffusion denoising. 
Simultaneously, we adaptively adjust the guidance strength based on confidence masks and diffusion time steps, enhancing the model's early-stage stability and convergence while preserving its ability to refine details during the later stages of diffusion.

\smallsec{Time Step Decay Scheduling. }
Diffusion models operate under different noise levels across time steps: early steps are highly noisy and emphasize coarse global structure, whereas later steps refine fine details. 
Accordingly, the guidance strength should be stronger when noise is large to stabilize and accelerate convergence, and should weaken as denoising progresses to avoid over-constraining detail refinement. 
We implement this with a time-decay weight as follows:
\begin{equation}
    W_t = (1.0 - t_{cur} / T) \cdot s,
\end{equation}
where $W_t$ represents the guidance weight of the current time step, $T$ represents the predefined total number of diffusion sampling steps, $t_{cur}$ denotes the current time step valued in \{T, T - 1, ..., 0\}, and $s$ is the fixed guidance scale for simple optical flow, representing the upper limit of guidance intensity.
Following FlowDiffuser \cite{luo2024flowdiffuser}, we set $T=6$ in both training and inference. We set $s=0.95$, slightly below 1, to prevent the simple-flow prior from becoming a hard constraint.
This schedule decays the guidance as the process moves toward later steps, yielding a smooth transition in how strongly the base prior influences the updates.

\smallsec{Confidence-Based Initialization. }
We first construct a simple-flow prior by weighting the rough flow $F_{r}$ with the confidence mask $M_{conf}$:
\begin{equation}
F_{s} = F_{r} \odot M_{conf},
\end{equation}
where $F_{s}$ represents the simple-flow prior and $\odot$ denotes pixel-wise multiplication.
Here, $F_s$ is not intended to correct angular errors by simply shrinking the flow magnitude. Instead, $M_{conf}$ serves as a reliability gate: high-confidence rough flow provides a structural prior, while low-confidence regions suppress the prior and rely more on image conditioning and diffusion refinement.
We then initialize the diffusion process by filling high-confidence regions with the simple-flow prior:
\begin{equation}
\begin{split}
B &= \mathbf{1}[M_{conf} > \tau], \\
x_T &= B \odot F_s + (1-B)\odot \epsilon,
\end{split}
\end{equation}
where $B$ denotes the binary initialization mask, $\tau=0.5$ is the confidence threshold, and $\epsilon \sim \mathcal{N}(0,I)$ denotes Gaussian noise with the same spatial size as the optical flow.
Unlike hard masked denoising, we do not clamp the initialized high-confidence regions during subsequent denoising steps. This allows the denoiser to further correct the whole flow field when image evidence suggests necessary refinements.

\smallsec{Confidence-Gated Residual Guidance. }
After initialization, we inject the simple-flow prior through residual-form guidance at each denoising step.
Given the current flow estimate $\hat{F}t$, we compute a confidence-gated residual:
\begin{equation}
\begin{split}
G_{t} &= W_{t} \cdot F_s, \\
R_t &= G_t - W_t \cdot M_{conf} \odot \hat{F}_t,
\end{split}
\end{equation}
where $G_t$ denotes the guiding flow and $R_t$ encourages the current estimate to remain close to the reliable simple-flow prior in high-confidence regions, while having limited influence in uncertain regions.
The residual guidance is then encoded by a lightweight convolutional encoder $\phi_g(\cdot)$ and fused into the denoiser feature by direct addition:
\begin{equation}
\tilde{H}_t = H_t + \phi_g(R_t),
\end{equation}
where $H_t$ denotes the denoiser feature at time step $t$, and $\tilde{H}_t$ denotes the guidance-enhanced feature.
This soft residual injection avoids repeatedly adding the full prior to the denoising state, and is more flexible than hard-clamping known regions. As $W_t$ decays over time steps, the effect of residual guidance gradually weakens, allowing the model to preserve early-stage stability while retaining late-stage flexibility for detail refinement.

\smallsec{Complementarity with Image Conditioning. }
While the simple-flow prior provides reliable motion structure in easy regions, optical flow estimation fundamentally depends on image content. 
We therefore retain the original feature extraction components used in diffusion-based flow models \cite{luo2024flowdiffuser} to extract features from the input frames. 
These features encode semantic cues and pixel-wise matching relationships and are crucial for handling challenging conditions such as occlusions, textureless regions, and motion boundaries. 
Importantly, the guidance mechanism and image features act in a complementary manner: in simple regions, the simple-flow prior provides a stable structural anchor while image features support local refinement; in hard regions, the confidence mask suppresses unreliable prior guidance and the model relies primarily on image-derived features and diffusion refinement. 
This cooperative design enables the network to adaptively balance prior-driven constraints and data-driven corrections, leading to more stable training and improved final accuracy.

\smallsec{Unified Design for Training and Inference. }
We implement the same guidance strategy consistently during both training and inference.
During training, one diffusion timestep is uniformly sampled for each image pair, and the corresponding confidence-gated residual guidance is computed and injected into the denoiser feature. This ensures that the denoising process is continually regularized by the simple-flow prior without imposing a hard constraint.
During inference, we use the same number of sampling steps as FlowDiffuser \cite{luo2024flowdiffuser}, i.e., $T=6$, and apply the identical confidence-based initialization and step-wise residual guidance.

\section{Experiments}
\label{sec:experiments}

\subsection{Implementation Details}
\label{sec:exp_details}

\begin{figure}[h]
    \centering
    \begin{overpic}[width=0.95\textwidth]{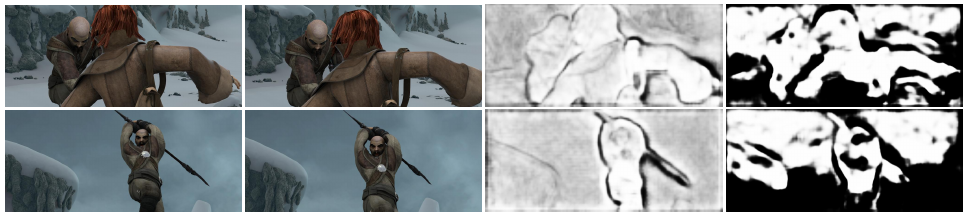}
    \put(8,23){Frame 1}
    \put(32,23){Frame 2}
        \put(49.5,23){Unsupervised results}
        \put(76.3,23){Supervised results}
    \end{overpic}
    \caption{Comparison of output results when training Confidence-Aware Network with or without explicit supervision of the confidence mask, where white denotes high confidence and black denotes low confidence.}
    \label{fig:result1}
\end{figure}
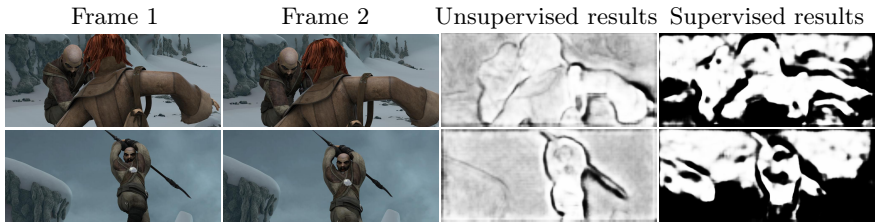

\smallsec{Network Architecture. }
\textit{1) Confidence-Aware Network: }
Our Confidence-Aware Network adopts a PWC-Net-style design \cite{sun2018pwc,zhao2020maskflownet}. 
We further incorporate several architectural adjustments inspired by MaskFlowNet \cite{zhao2020maskflownet}. 
Unlike MaskFlowNet \cite{zhao2020maskflownet}, which learns an occlusion mask in an unsupervised manner, we introduce explicit supervision for the confidence mask. 
This is crucial because our mask is expected to reflect not only occluded regions but also hard-flow regions. 
As shown in \figref{fig:result1}, explicit supervision helps the network more clearly distinguish regions where the rough flow is reliable, whereas unsupervised learning tends to produce overly conservative, medium-confidence predictions. 
Such more faithful confidence estimation provides a more informative reliability prior for the downstream inpainting-based refinement. 
\textit{2) Flow Inpainting Network: }
Given the strong performance and flow-specific design of FlowDiffuser \cite{luo2024flowdiffuser}, we adopt FlowDiffuser \cite{luo2024flowdiffuser} as the backbone for the diffusion component. 
Based on this backbone, we modify the initialization strategy and introduce a time step decay module to inject additional conditioning from the simple-flow prior. 
This reformulates the diffusion process from full dense-flow generation into a soft inpainting-style refinement process, where the model is guided by reliable flow in simple regions and mainly refines uncertain regions.

\smallsec{Training Details. }
\textit{1) Training of Confidence-Aware Network. }
We train the Confidence-Aware Network in two stages. 
In the first stage, we pre-train on FlyingChairs \cite{dosovitskiy2015flownet} and FlyingThings3D \cite{mayer2016large}, using supervision only for flow prediction. 
The loss consists of an EPE loss and a self-supervised warping loss, weighted by 0.1 and 0.01, respectively. 
This stage equips the network with the ability to produce stable and reasonable flow estimates. 
In the second stage, we fine-tune on Sintel \cite{Butler:ECCV:2012} (S+T) and KITTI \cite{Menze2015CVPR} while introducing supervision for the confidence mask. 
The mask construction is described in \secref{sec:method_stage1}. We supervise the confidence mask using $L_{1}$ loss and set its weight to 0.8.
\textit{2) Training of Flow Inpainting Network.}
Following prior optical flow pipelines \cite{teed2020raft,luo2024flowdiffuser}, FlowPainter is trained sequentially on FlyingChairs \cite{dosovitskiy2015flownet}, FlyingThings3D \cite{mayer2016large}, Sintel \cite{Butler:ECCV:2012}, and KITTI \cite{Menze2015CVPR}. 
Although SEA-RAFT \cite{wang2024sea} shows the benefit of large-scale rigid-flow pretraining for iterative refinement paradigms, our current setting does not use such pretraining.
We empirically find that the confidence-aware prior and the diffusion backbone together enable competitive performance under this training protocol. 
A practical benefit of introducing the confidence-aware prior is that the diffusion stage requires substantially fewer training iterations than FlowDiffuser \cite{luo2024flowdiffuser}. 
Under the same batch size setting, FlowDiffuser \cite{luo2024flowdiffuser} typically needs 100k/200k/180k/50k iterations to converge on FlyingChairs \cite{dosovitskiy2015flownet}/FlyingThings3D \cite{mayer2016large}/Sintel \cite{Butler:ECCV:2012}/KITTI \cite{Menze2015CVPR}, respectively. 
In contrast, FlowPainter converges with 20k/50k/50k/15k iterations on the same datasets. 
With 8$\times$A100 GPUs, this reduces the diffusion-stage training time from 7 days 2 hours to 2 days 1 hour, while also improving the final accuracy.

\begin{table}[h]
  \centering
  \caption{Quantitative comparison of FlowPainter with state-of-the-art optical flow methods on the train splits. The best results are \textbf{highlighted in bold. }}
  \scalebox{0.99}{
    \begin{tabular}{cccccccc}
    \toprule
    \multirow{2}[4]{*}{Method} & \multirow{2}[4]{*}{Extra Data} & \multicolumn{2}{c}{Sintel(train)} & \multicolumn{2}{c}{KITTI-15(train)} & \multicolumn{2}{c}{Spring(train)} \\
\cmidrule(r){3-4} \cmidrule{5-6} \cmidrule(l){7-8}         &       & Clean & Final & EPE   & Fl-all & \multicolumn{2}{c}{EPE} \\
    \midrule
    SEA-RAFT \cite{wang2024sea} & Tartan, Spring & -     & -     & -     & -     & \multicolumn{2}{c}{0.41} \\
    FlowDiffuser \cite{luo2024flowdiffuser} & Spring & -     & -     & -     & -     & \multicolumn{2}{c}{0.43} \\
    FlowPainter(Ours) & Spring & -     & -     & -     & -     & \multicolumn{2}{c}{\textbf{0.40}} \\
    \midrule
    RAFT \cite{teed2020raft}  & -     & 1.43  & 2.71  & 5.04  & 17.4  & \multicolumn{2}{c}{0.45} \\
    GMA \cite{jiang2021learning}   & -     & 1.30   & 2.74  & 4.69  & 17.1  & \multicolumn{2}{c}{0.44} \\
    GMFlow \cite{xu2022gmflow} & -     & 1.08  & 2.48  & 7.75  & 23.2  & \multicolumn{2}{c}{0.93} \\
    FlowFormer \cite{huang2022flowformer} & -     & 0.99  & 2.38  & 4.11  & 14.6  & \multicolumn{2}{c}{0.47} \\
    GAFlow \cite{luo2023gaflow} & -     & 0.95  & 2.33  & 3.92  & 13.9  & \multicolumn{2}{c}{0.46} \\
    GMFlow+ \cite{xu2023unifying} & -     & 0.91  & 2.74  & 5.74  & 17.6  & \multicolumn{2}{c}{0.43} \\
    FlowFormer++ \cite{shi2023flowformer++} & YouTube-VOS & 0.90   & 2.30   & 3.93  & 14.1  & \multicolumn{2}{c}{0.45} \\
    MatchFlow \cite{dong2023rethinking} & MegaDepth & 1.03  & 2.45  & 4.08  & 15.6  & \multicolumn{2}{c}{0.41} \\
    SEA-RAFT \cite{wang2024sea} & Tartan & 1.18  & 4.13  & 3.62  & 12.8  & \multicolumn{2}{c}{-} \\
    FlowSeek \cite{poggi2025flowseek} & Tartan & 1.04  & 2.18  & 3.34  & 11.3  & \multicolumn{2}{c}{\textbf{0.40}} \\
    FlowDiffuser \cite{luo2024flowdiffuser} & -     & 0.88  & 2.23  & 3.75  & 11.1  & \multicolumn{2}{c}{-} \\
    FlowPainter(Ours) & -     & \textbf{0.87} & \textbf{1.32} & \textbf{3.17} & \textbf{6.84} & \multicolumn{2}{c}{-} \\
\bottomrule    \end{tabular}%
    }
  \label{tab:compare_main_train}%
\end{table}%

\begin{table}[htbp]
  \centering
  \setlength{\tabcolsep}{12pt}
  \caption{Quantitative comparison of FlowPainter with state-of-the-art optical flow methods on the test splits. The best results are \textbf{highlighted in bold. }}
  \vspace{5mm}
    \begin{tabular}{cccccc}
    \toprule
    \multirow{2}[4]{*}{Method} & \multicolumn{2}{c}{Sintel(test)} & \multicolumn{3}{c}{KITTI-15(test)} \\
\cmidrule(r){2-3} \cmidrule(l){4-6}          & Clean & Final & Fl-bg & Fl-fg & Fl-all \\
    \midrule
    RAFT \cite{teed2020raft}  & 1.61  & 2.86  & 4.74  & 6.87  & 5.10  \\
    GMFlow \cite{xu2022gmflow} & 1.74  & 2.90  & 9.67  & 7.57  & 9.32  \\
    FlowFormer \cite{huang2022flowformer} & 1.16  & 2.09  & 4.37  & 6.18  & 4.68  \\
    GMFlow+ \cite{xu2023unifying} & 1.03  & 2.37  & 4.27  & 5.60  & 4.49  \\
    MatchFlow \cite{dong2023rethinking} & 1.16  & 2.37  & 4.33  & 6.11  & 4.63  \\
    SEA-RAFT \cite{wang2024sea} & 1.31  & 2.60  & -     & -     & - \\
    FlowDiffuser \cite{luo2024flowdiffuser} & 1.02  & 2.03  & 3.68  & 6.64  & 4.17  \\
    DPFlow \cite{morimitsu2025dpflow} & 1.05  & 1.98  & 3.29  & \textbf{4.93} & 3.56  \\
    MemFlow \cite{dong2024memflow} & 1.05  & 1.91  & 3.67  & 6.27  & 4.10  \\
    MegaFlow \cite{zhang2026megaflow} & 1.49  & 2.43  & 3.55  & 5.89  & 3.94  \\
    DDVM \cite{saxena2023surprising}  & 1.75  & 2.48  & 2.90  & 5.05  & 3.26  \\
    FlowPainter(Ours) & \textbf{1.01} & \textbf{1.71} & \textbf{2.35} & 6.37 & \textbf{3.02} \\
    \bottomrule
    \end{tabular}%
  \label{tab:compare_main_test}%
\end{table}%

\subsection{Evaluation Results}
\label{sec:metrics}

\smallsec{Quantitative Evaluation. }
We conduct quantitative evaluations on Sintel \cite{Butler:ECCV:2012}, KITTI \cite{Menze2015CVPR} and Spring \cite{Mehl2023_Spring} benchmarks. The results are summarized in \tabref{tab:compare_main_train} and \tabref{tab:compare_main_test}. 
Under comparable training settings, FlowPainter achieves the best results on most major metrics in the listed comparison, while using no additional training data in the Sintel \cite{Butler:ECCV:2012} and KITTI \cite{Menze2015CVPR} setting. 
On Sintel(train), FlowPainter achieves 0.87/1.32 EPE on the clean/final passes, and also obtains strong results on Sintel(test) with 1.01/1.71 EPE. 
Compared with the diffusion-based baseline FlowDiffuser \cite{luo2024flowdiffuser}, the improvement is modest on the clean pass (1\% on train; 1\% on test), but becomes more pronounced on the final pass (40\% on train; 16\% on test). 
Since the Sintel final pass contains degradations such as motion blur, defocus, atmospheric effects, and complex illumination changes, this larger improvement suggests that FlowPainter is more robust under the combined challenging conditions represented by this benchmark split, without implying isolated factor-wise verification for each degradation type. 
On KITTI benchmark, FlowPainter also improves over FlowDiffuser \cite{luo2024flowdiffuser}, reducing EPE by 15\% and Fl-all by 39\% on the train split relative to FlowDiffuser \cite{luo2024flowdiffuser}. On the test split, FlowPainter also achieves the best Fl-all results among the listed methods. These results suggest that FlowPainter generalizes well beyond Sintel and remains competitive in real-world driving scenarios such as occlusions, large displacements, and thin structures. 
For Spring \cite{Mehl2023_Spring}, we additionally report results under Spring-based training: FlowPainter achieves 0.40 EPE, outperforming both SEA-RAFT \cite{wang2024sea} (0.41 EPE) and FlowDiffuser \cite{luo2024flowdiffuser} (0.43 EPE), suggesting that the proposed confidence-guided inpainting formulation is also effective under different motion statistics.

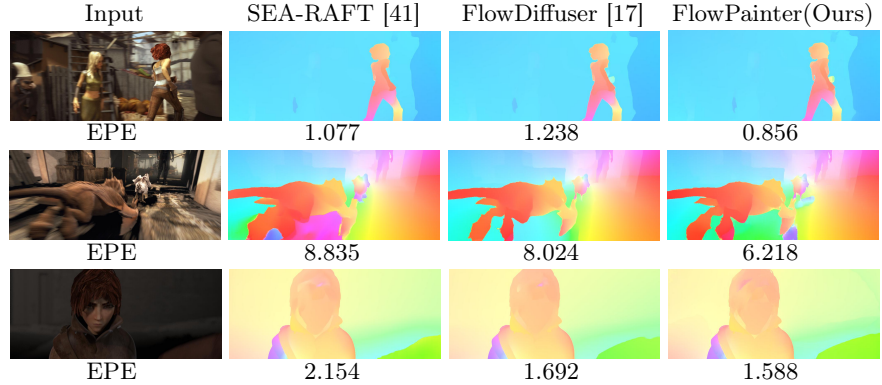
\begin{figure}[h]
    \centering
    \begin{overpic}[width=0.95\textwidth]{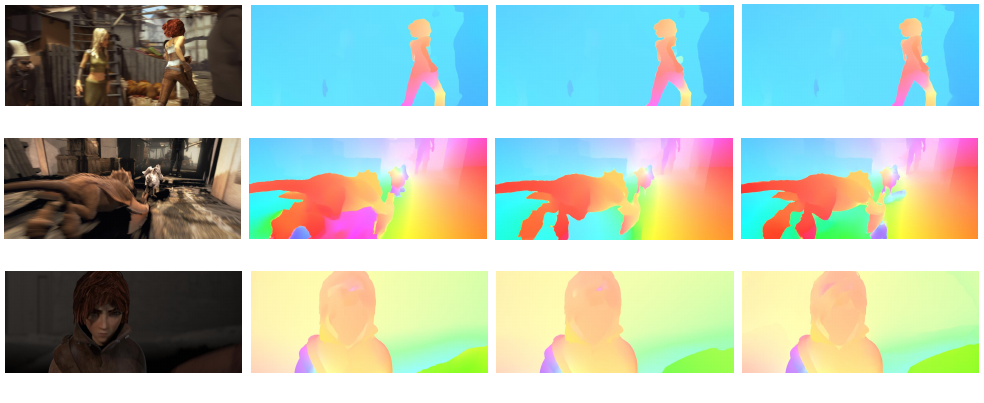}
    \put(9,41.5){Input}
    \put(27.5,41.5){SEA-RAFT \cite{wang2024sea}}
    \put(52,41.5){FlowDiffuser \cite{luo2024flowdiffuser}}
    \put(76,41.5){FlowPainter(Ours)}
    
    \put(9.2,28){EPE}
    \put(34,28){1.077}
    \put(59,28){1.238}
    \put(84,28){0.856}
        
    \put(9.2,14.3){EPE}
    \put(34,14.3){8.835}
    \put(59,14.3){8.024}
    \put(84,14.3){6.218}
    
    \put(9.2,0.8){EPE}
    \put(34,0.8){2.154}
    \put(59,0.8){1.692}
    \put(84,0.8){1.588}
    \end{overpic}
    \caption{Qualitative comparison of FlowPainter with SEA-RAFT \cite{wang2024sea} and FlowDiffuser \cite{luo2024flowdiffuser} on Sintel(test). }
    \label{fig:compare_main}
\end{figure}

\smallsec{Qualitative Evaluation. }
As illustrated in \figref{fig:compare_main}, we compare flow visualizations of FlowPainter on Sintel(test) \cite{Butler:ECCV:2012} against SEA-RAFT \cite{wang2024sea} and FlowDiffuser \cite{luo2024flowdiffuser}. 
The examples show that diffusion-based methods can produce detailed flow estimates in complex scenes and handle irregular motions more effectively than purely discriminative iterative refinement in some cases. 
FlowPainter further improves the visual quality by using reliable simple-flow priors while preserving the refinement capability of the diffusion backbone in hard regions.

\subsection{Ablation Studies}
\label{ablation}

\smallsec{Confidence Mask Construction Strategy. }
The reliability of the confidence mask predicted by the Confidence-Aware Network is crucial for downstream inpainting, as it determines where the diffusion model should rely more on image conditioning and denoising refinement.
We therefore ablate the construction of the confidence mask during training.
As shown in \tabref{tab:ablation1}, both the occlusion mask and the hard-flow mask contribute to the final performance.
Discarding either signal leads to a clear drop in accuracy, suggesting that they capture complementary sources of unreliability: occlusions violate the correspondence assumptions underlying flow estimation, whereas hard-flow regions are difficult for the lightweight base network to estimate reliably.
Explicitly supervising both aspects yields a sharper and more informative confidence mask, which in turn enables more effective guidance and region-adaptive refinement.

\begin{minipage}[c]{0.45\textwidth}
		\centering
		\captionof{table}{Ablation on occlusion mask and hard-flow mask, where O.M. denotes occlusion mask and H.M. denotes hard-flow mask. }
		\setlength{\tabcolsep}{0.5 mm}
		\scalebox{0.95}{
		\begin{tabular}{cccc}
    \toprule
    \multirow{2}[4]{*}{O.M.} & \multirow{2}[4]{*}{H.M.} & Sintel(train) & KITTI(train) \\
\cmidrule{3-4}          &       & Clean/Final & EPE/Fl-all \\
    \midrule
    \usym{2613}     & \usym{2613}     & 1.05/2.17 & 3.69/8.92 \\
    \usym{2613}     & \checkmark     & 0.88/1.37 & 3.22/7.01 \\
    \checkmark     & \usym{2613}     & 1.01/1.95 & 3.47/8.13 \\
    \checkmark     & \checkmark     & \textbf{0.87/1.32} & \textbf{3.17/6.84} \\
    \bottomrule
    \end{tabular}
	}
    \label{tab:ablation1}
    \vspace{5mm}
\end{minipage}
\hfill
\begin{minipage}[c]{0.45\textwidth}
		\centering
		\captionof{table}{Ablation on time step decay module and inpainting strategy, where T.D. denotes time step decay module and I.P. denotes inpainting strategy. }
		\setlength{\tabcolsep}{0.5mm}
		\scalebox{0.95}{
		\begin{tabular}{cccc}
    \toprule
    \multirow{2}[4]{*}{T.D.} & \multirow{2}[4]{*}{I.P.} & Sintel(train) & KITTI(train) \\
\cmidrule{3-4}          &       & Clean/Final & EPE/Fl-all \\
    \midrule
    \usym{2613}     & \usym{2613}     & 1.23/2.74 & 4.52/12.18 \\
    \usym{2613}     & \checkmark     & 1.11/2.32 & 3.98/10.12 \\
    \checkmark     & \usym{2613}     & 0.95/1.67 & 3.49/8.97 \\
    \checkmark     & \checkmark     & \textbf{0.87/1.32} & \textbf{3.17/6.84} \\
    \bottomrule
    \end{tabular}
	}
    \label{tab:ablation2}
    \vspace{5mm}
\end{minipage}

\smallsec{Time Step Decay Module and Inpainting Strategy. }
We conduct ablations on the time step decay module and the inpainting formulation within the Flow Inpainting Network. Here, removing the inpainting strategy means removing the confidence-mask-based inpainting guidance.
As shown in \tabref{tab:ablation2}, removing either component consistently degrades performance, showing that the two components are complementary in the proposed refinement pipeline.
In particular, the time step decay module provides a schedule for controlling the strength of the simple-flow prior across diffusion steps, which stabilizes optimization in the high-noise regime while avoiding over-constraint during late-stage detail refinement.
Meanwhile, the inpainting formulation reduces the denoising burden in reliable simple regions and biases refinement toward uncertain hard regions, leading to more accurate and robust estimates especially in challenging cases.

\begin{table}[htbp]
  \centering
  \caption{Ablation study result for simple flow initialization thresholds. }
    \setlength{\tabcolsep}{8pt}
    \begin{tabular}{ccccccc}
    \toprule
    \multicolumn{2}{c}{Threshold} & 0.3   & 0.4   & 0.5   & 0.6   & 0.7 \\
    \midrule
    \multirow{2}[2]{*}{Sintel(train)} & Clean & 0.95  & 0.88  & \textbf{0.87} & 0.91  & 0.97 \\
          & Final & 1.78  & 1.57  & \textbf{1.32} & 1.63  & 1.85 \\
    \midrule
    \multirow{2}[2]{*}{KITTI(train)} & EPE   & 3.76  & 3.37  & \textbf{3.17} & 3.35  & 3.98 \\
          & Fl-all & 8.54  & 7.93  & \textbf{6.84} & 7.88  & 8.71 \\
    \bottomrule
    \end{tabular}%
  \label{tab:ablation_threshold}%
\end{table}%

\smallsec{Initialization Threshold for Simple Flow. }
When the confidence value exceeds a preset threshold, the corresponding simple-flow prediction is used to initialize the diffusion process.
This threshold controls the trade-off between the quantity and reliability of injected prior information.
If the threshold is set too low, inaccurate base-flow estimates may be introduced into the initialization, increasing the correction burden and potentially biasing refinement in uncertain regions.
Conversely, an overly conservative threshold limits the usable prior, weakening the benefit of guided inpainting and leading to slower convergence and reduced accuracy.
We evaluate thresholds {0.3, 0.4, 0.5, 0.6, 0.7}; the results in \tabref{tab:ablation_threshold} show that 0.5 achieves the best trade-off among the tested thresholds.

\section{Conclusion}
\label{sec:conclusion}

We proposed \textit{FlowPainter}, a confidence-guided diffusion framework that integrates reliable lightweight flow priors with diffusion-based refinement for optical flow estimation.
By using confidence-based initialization and time-decayed residual guidance, FlowPainter reformulates dense-flow generation from noise as a soft inpainting-style refinement problem, reducing the denoising burden in reliable regions while preserving flexibility for uncertain hard regions.
Experiments on Sintel, KITTI, and Spring show that FlowPainter achieves strong accuracy under comparable training settings and improves convergence efficiency over existing diffusion-based optical flow methods.

\bibliographystyle{splncs04}
\bibliography{main}

@String(PAMI  = {IEEE Trans. Pattern Anal. Mach. Intell.})

@String(IJCV  = {Int. J. Comput. Vis.})

@String(CVPR  = {IEEE Conf. Comput. Vis. Pattern Recog.})

@String(ICCV  = {Int. Conf. Comput. Vis.})

@String(ECCV  = {Eur. Conf. Comput. Vis.})

@String(NeurIPS = {Adv. Neural Inform. Process. Syst.})

@String(TOG   = {ACM Trans. Graph.})

@String(CVM   = {Computational Visual Media})

@String(PAMI  = {IEEE TPAMI})

@String(IJCV  = {IJCV})

@String(CVPR  = {CVPR})

@String(ICCV  = {ICCV})

@String(ECCV  = {ECCV})

@String(NeurIPS = {NeurIPS})

@String(TOG   = {ACM TOG})

@inproceedings{Butler:ECCV:2012,
title = {A naturalistic open source movie for optical flow evaluation},
author = {Butler, D. J. and Wulff, J. and Stanley, G. B. and Black, M. J.},
booktitle = ECCV,
pages = {611--625},
year = {2012}
}

@inproceedings{Wulff:ECCVws:2012,
 title = {Lessons and insights from creating a synthetic optical flow benchmark},
 author = {Wulff, J. and Butler, D. J. and Stanley, G. B. and Black, M. J.},
 booktitle = ECCV,
 pages = {168--177},
 year = {2012}
}

@inproceedings{Menze2015CVPR,
  author = {Moritz Menze and Andreas Geiger},
  title = {Object Scene Flow for Autonomous Vehicles},
  booktitle = CVPR,
  year = {2015}
}

@inproceedings{luo2024flowdiffuser,
  title={FlowDiffuser: Advancing Optical Flow Estimation with Diffusion Models},
  author={Luo, Ao and Li, Xin and Yang, Fan and Liu, Jiangyu and Fan, Haoqiang and Liu, Shuaicheng},
  booktitle=CVPR,
  pages={19167--19176},
  year={2024}
}

@inproceedings{wang2024sea,
  title={Sea-raft: Simple, efficient, accurate raft for optical flow},
  author={Wang, Yihan and Lipson, Lahav and Deng, Jia},
  booktitle=ECCV,
  pages={36--54},
  year={2024}
}

@inproceedings{liu2025flow4agent,
  title={Flow4agent: Long-form video understanding via motion prior from optical flow},
  author={Liu, Ruyang and Sun, Shangkun and Tang, Haoran and Gao, Wei and Li, Ge},
  booktitle=ICCV,
  pages={23817--23827},
  year={2025}
}

@inproceedings{fan2018end,
  title={End-to-end learning of motion representation for video understanding},
  author={Fan, Lijie and Huang, Wenbing and Gan, Chuang and Ermon, Stefano and Gong, Boqing and Huang, Junzhou},
  booktitle=CVPR,
  pages={6016--6025},
  year={2018}
}

@inproceedings{teed2020raft,
  title={Raft: Recurrent all-pairs field transforms for optical flow},
  author={Teed, Zachary and Deng, Jia},
  booktitle=ECCV,
  pages={402--419},
  year={2020}
}

@inproceedings{piergiovanni2019representation,
  title={Representation flow for action recognition},
  author={Piergiovanni, AJ and Ryoo, Michael S},
  booktitle=CVPR,
  pages={9945--9953},
  year={2019}
}

@inproceedings{sun2018optical,
  title={Optical flow guided feature: A fast and robust motion representation for video action recognition},
  author={Sun, Shuyang and Kuang, Zhanghui and Sheng, Lu and Ouyang, Wanli and Zhang, Wei},
  booktitle=CVPR,
  pages={1390--1399},
  year={2018}
}

@article{shen2023optical,
  title={Optical flow for autonomous driving: Applications, challenges and improvements},
  author={Shen, Shihao and Kerofsky, Louis and Yogamani, Senthil},
  journal={arXiv preprint arXiv:2301.04422},
  year={2023}
}

@article{mahjourian2022occupancy,
  title={Occupancy flow fields for motion forecasting in autonomous driving},
  author={Mahjourian, Reza and Kim, Jinkyu and Chai, Yuning and Tan, Mingxing and Sapp, Ben and Anguelov, Dragomir},
  journal={RA-L},
  volume={7},
  pages={5639--5646},
  year={2022},
  publisher={IEEE}
}

@inproceedings{capito2020optical,
  title={Optical flow based visual potential field for autonomous driving},
  author={Capito, Linda and Ozguner, Umit and Redmill, Keith},
  booktitle={IV},
  pages={885--891},
  year={2020}
}

@inproceedings{raket2012motion,
  title={Motion compensated frame interpolation with a symmetric optical flow constraint},
  author={Rak{\^e}t, Lars Lau and Roholm, Lars and Bruhn, Andr{\'e}s and Weickert, Joachim},
  booktitle={ISVC},
  pages={447--457},
  year={2012}
}

@inproceedings{wu2022video,
  title={Video interpolation by event-driven anisotropic adjustment of optical flow},
  author={Wu, Song and You, Kaichao and He, Weihua and Yang, Chen and Tian, Yang and Wang, Yaoyuan and Zhang, Ziyang and Liao, Jianxing},
  booktitle=ECCV,
  pages={267--283},
  year={2022}
}

@inproceedings{hai2025hierarchical,
  title={Hierarchical flow diffusion for efficient frame interpolation},
  author={Hai, Yang and Wang, Guo and Su, Tan and Jiang, Wenjie and Hu, Yinlin},
  booktitle=CVPR,
  pages={22943--22952},
  year={2025}
}

@inproceedings{dosovitskiy2015flownet,
  title={Flownet: Learning optical flow with convolutional networks},
  author={Dosovitskiy, Alexey and Fischer, Philipp and Ilg, Eddy and Hausser, Philip and Hazirbas, Caner and Golkov, Vladimir and Van Der Smagt, Patrick and Cremers, Daniel and Brox, Thomas},
  booktitle=ICCV,
  pages={2758--2766},
  year={2015}
}

@inproceedings{ilg2017flownet,
  title={Flownet 2.0: Evolution of optical flow estimation with deep networks},
  author={Ilg, Eddy and Mayer, Nikolaus and Saikia, Tonmoy and Keuper, Margret and Dosovitskiy, Alexey and Brox, Thomas},
  booktitle=CVPR,
  pages={2462--2470},
  year={2017}
}

@inproceedings{ranjan2017optical,
  title={Optical flow estimation using a spatial pyramid network},
  author={Ranjan, Anurag and Black, Michael J},
  booktitle=CVPR,
  pages={4161--4170},
  year={2017}
}

@inproceedings{sun2018pwc,
  title={Pwc-net: Cnns for optical flow using pyramid, warping, and cost volume},
  author={Sun, Deqing and Yang, Xiaodong and Liu, Ming-Yu and Kautz, Jan},
  booktitle=CVPR,
  pages={8934--8943},
  year={2018}
}

@inproceedings{zhao2020maskflownet,
  title={Maskflownet: Asymmetric feature matching with learnable occlusion mask},
  author={Zhao, Shengyu and Sheng, Yilun and Dong, Yue and Chang, Eric I and Xu, Yan and others},
  booktitle=CVPR,
  pages={6278--6287},
  year={2020}
}

@inproceedings{jiang2021learning,
  title={Learning to estimate hidden motions with global motion aggregation},
  author={Jiang, Shihao and Campbell, Dylan and Lu, Yao and Li, Hongdong and Hartley, Richard},
  booktitle=ICCV,
  pages={9772--9781},
  year={2021}
}

@inproceedings{teed2021raft,
  title={Raft-3d: Scene flow using rigid-motion embeddings},
  author={Teed, Zachary and Deng, Jia},
  booktitle=CVPR,
  pages={8375--8384},
  year={2021}
}

@inproceedings{poggi2025flowseek,
  title={FlowSeek: optical flow made easier with depth foundation models and motion bases},
  author={Poggi, Matteo and Tosi, Fabio},
  booktitle=ICCV,
  pages={5667--5679},
  year={2025}
}

@inproceedings{saxena2023surprising,
  title={The surprising effectiveness of diffusion models for optical flow and monocular depth estimation},
  author={Saxena, Saurabh and Herrmann, Charles and Hur, Junhwa and Kar, Abhishek and Norouzi, Mohammad and Sun, Deqing and Fleet, David J},
  booktitle=NeurIPS,
  pages={39443--39469},
  year={2023}
}

@article{dong2023open,
  title={Open-ddvm: A reproduction and extension of diffusion model for optical flow estimation},
  author={Dong, Qiaole and Zhao, Bo and Fu, Yanwei},
  journal={arXiv preprint arXiv:2312.01746},
  year={2023}
}

@article{zhao2026resilphaseplugandplayphasemapping,
  title={ResilPhase: Plug-and-Play Phase Mapping and Noise-Resilient Macro-Trajectory Extrapolation for Diffusion Acceleration},
  author={Qicheng Zhao and Yu Li and Qi Sun and Zheyu Yan},
  journal={arXiv preprint arXiv:2606.26769},
  year={2026}
}

@inproceedings{pepe2025geometric,
  title={Geometric Inductive Priors in Diffusion-Based Optical Flow Estimation},
  author={Pepe, Alberto and Dos Santos Mendonca, Paulo and Lasenby, Joan},
  booktitle=ICCV,
  pages={655--665},
  year={2025}
}

@InProceedings{Mehl2023_Spring,
    author    = {Lukas Mehl and Jenny Schmalfuss and Azin Jahedi and Yaroslava Nalivayko and Andr\'es Bruhn},
    title     = {Spring: A High-Resolution High-Detail Dataset and Benchmark for Scene Flow, Optical Flow and Stereo},
    booktitle =CVPR,
    year      = {2023}
}

@inproceedings{xu2022gmflow,
  title={Gmflow: Learning optical flow via global matching},
  author={Xu, Haofei and Zhang, Jing and Cai, Jianfei and Rezatofighi, Hamid and Tao, Dacheng},
  booktitle=CVPR,
  pages={8121--8130},
  year={2022}
}

@inproceedings{huang2022flowformer,
  title={Flowformer: A transformer architecture for optical flow},
  author={Huang, Zhaoyang and Shi, Xiaoyu and Zhang, Chao and Wang, Qiang and Cheung, Ka Chun and Qin, Hongwei and Dai, Jifeng and Li, Hongsheng},
  booktitle=ECCV,
  pages={668--685},
  year={2022},
  organization={Springer}
}

@inproceedings{gao2023implicit,
  title={Implicit diffusion models for continuous super-resolution},
  author={Gao, Sicheng and Liu, Xuhui and Zeng, Bohan and Xu, Sheng and Li, Yanjing and Luo, Xiaoyan and Liu, Jianzhuang and Zhen, Xiantong and Zhang, Baochang},
  booktitle=CVPR,
  pages={10021--10030},
  year={2023}
}

@inproceedings{zhang2023adding,
  title={Adding conditional control to text-to-image diffusion models},
  author={Zhang, Lvmin and Rao, Anyi and Agrawala, Maneesh},
  booktitle=ICCV,
  pages={3836--3847},
  year={2023}
}

@article{zhang2024clay,
  title={Clay: A controllable large-scale generative model for creating high-quality 3d assets},
  author={Zhang, Longwen and Wang, Ziyu and Zhang, Qixuan and Qiu, Qiwei and Pang, Anqi and Jiang, Haoran and Yang, Wei and Xu, Lan and Yu, Jingyi},
  journal={TOG},
  volume={43},
  number={4},
  pages={1--20},
  year={2024},
  publisher={ACM New York, NY, USA}
}

@article{wang2024exploiting,
  title={Exploiting diffusion prior for real-world image super-resolution},
  author={Wang, Jianyi and Yue, Zongsheng and Zhou, Shangchen and Chan, Kelvin CK and Loy, Chen Change},
  journal=IJCV,
  volume={132},
  number={12},
  pages={5929--5949},
  year={2024},
  publisher={Springer}
}

@inproceedings{ho2020denoising,
  title={Denoising diffusion probabilistic models},
  author={Ho, Jonathan and Jain, Ajay and Abbeel, Pieter},
  booktitle=NeurIPS,
  pages={6840--6851},
  year={2020}
}

@article{song2020denoising,
  title={Denoising diffusion implicit models},
  author={Song, Jiaming and Meng, Chenlin and Ermon, Stefano},
  journal={arXiv preprint arXiv:2010.02502},
  year={2020}
}

@inproceedings{ji2023ddp,
  title={Ddp: Diffusion model for dense visual prediction},
  author={Ji, Yuanfeng and Chen, Zhe and Xie, Enze and Hong, Lanqing and Liu, Xihui and Liu, Zhaoqiang and Lu, Tong and Li, Zhenguo and Luo, Ping},
  booktitle=ICCV,
  pages={21741--21752},
  year={2023}
}

@inproceedings{mayer2016large,
  title={A large dataset to train convolutional networks for disparity, optical flow, and scene flow estimation},
  author={Mayer, Nikolaus and Ilg, Eddy and Hausser, Philip and Fischer, Philipp and Cremers, Daniel and Dosovitskiy, Alexey and Brox, Thomas},
  booktitle=CVPR,
  pages={4040--4048},
  year={2016}
}

@inproceedings{luo2023gaflow,
  title={Gaflow: Incorporating gaussian attention into optical flow},
  author={Luo, Ao and Yang, Fan and Li, Xin and Nie, Lang and Lin, Chunyu and Fan, Haoqiang and Liu, Shuaicheng},
  booktitle=ICCV,
  pages={9642--9651},
  year={2023}
}

@inproceedings{Zhang2021SepFlow,
  title={Separable Flow: Learning Motion Cost Volumes for Optical Flow Estimation},
  author={Zhang, Feihu and Woodford, Oliver J. and Prisacariu, Victor Adrian and Torr, Philip H.S.},
  booktitle=ICCV,
  year={2021},
  pages={10807-10817}
}

@inproceedings{dong2023rethinking,
  title={Rethinking Optical Flow from Geometric Matching Consistent  Perspective},
  author={Dong, Qiaole and Cao, Chenjie and Fu, Yanwei},
  booktitle=CVPR,
  year={2023}
}

@article{xu2023unifying,
  title={Unifying flow, stereo and depth estimation},
  author={Xu, Haofei and Zhang, Jing and Cai, Jianfei and Rezatofighi, Hamid and Yu, Fisher and Tao, Dacheng and Geiger, Andreas},
  journal=PAMI,
  volume={45},
  number={11},
  pages={13941--13958},
  year={2023},
  publisher={IEEE}
}

@inproceedings{shi2023flowformer++,
  title={Flowformer++: Masked cost volume autoencoding for pretraining optical flow estimation},
  author={Shi, Xiaoyu and Huang, Zhaoyang and Li, Dasong and Zhang, Manyuan and Cheung, Ka Chun and See, Simon and Qin, Hongwei and Dai, Jifeng and Li, Hongsheng},
  booktitle=CVPR,
  pages={1599--1610},
  year={2023}
}

@inproceedings{morimitsu2025dpflow,
  title={Dpflow: Adaptive optical flow estimation with a dual-pyramid framework},
  author={Morimitsu, Henrique and Zhu, Xiaobin and Cesar, Roberto M and Ji, Xiangyang and Yin, Xu-Cheng},
  booktitle=CVPR,
  pages={17810--17820},
  year={2025}
}

@article{wang2025diffusion,
  title={Diffusion models for 3D generation: A survey},
  author={Wang, Chen and Peng, Hao-Yang and Liu, Ying-Tian and Gu, Jiatao and Hu, Shi-Min},
  journal=CVM,
  volume={11},
  number={1},
  pages={1--28},
  year={2025},
  publisher={TUP}
}

@inproceedings{dong2024memflow,
  title={Memflow: Optical flow estimation and prediction with memory},
  author={Dong, Qiaole and Fu, Yanwei},
  booktitle=CVPR,
  pages={19068--19078},
  year={2024}
}

@article{zhang2026megaflow,
  title={MegaFlow: Zero-Shot Large Displacement Optical Flow},
  author={Zhang, Dingxi and Wang, Fangjinhua and Pollefeys, Marc and Xu, Haofei},
  journal={arXiv preprint arXiv:2603.25739},
  year={2026}
}

@article{chen2026calibrated,
  title={Calibrated Harmonic Overlaid Implicit Neural Representations for Multi-Dimensional Data},
  author={Chen, Honghang and Zhang, Xiujun and Sun, Xiaoli and Xiao Mingqing},
  journal={arXiv preprint arXiv:2606.26763},
  year={2026}
}

@inproceedings{meng2025ultraled,
  title={Ultraled: Learning to see everything in ultra-high dynamic range scenes},
  author={Meng, Yuang and Jin, Xin and Lei, Lina and Guo, Chun-Le and Li, Chongyi},
  booktitle=NeurIPS,
  pages={41466--41495},
  year={2025}
}

@article{qu2026there,
  title={There and Back Again: A Flexible-Frame Transformer for Multi-Exposure Fusion},
  author={Qu, Lishen and Liu, Yao and Zhou, Shihao and Liang, Jie and Zeng, Hui and Zhang, Lei and Yang, Jufeng},
  journal={arXiv preprint arXiv:2606.27905},
  year={2026}
}
\end{document}